# FashionBrain Project: A Vision for Understanding Europe's Fashion Data Universe

Alessandro Checco+, Gianluca Demartini+, Alexander Löser*, Ines Arous‡, Mourad Khayati‡,

Matthias Dantone†, Richard Koopmanschap§, Svetlin Stalinov§, Martin Kersten§, Ying Zhang§

+University of Sheffield, UK
*Beuth Hochschule für Technik Berlin, Germany
‡University of Fribourg, Switzerland
†Fashwell AG, Switzerland
§MonetDB Solutions, The Netherlands

## ABSTRACT

A core business in the fashion industry is the understanding and prediction of customer needs and trends. Search engines and social networks are at the same time a fundamental bridge and a costly middleman between the customer's purchase intention and the retailer. To better exploit Europe's distinctive characteristics e.g., multiple languages, fashion and cultural differences, it is pivotal to reduce retailers' dependence to search engines. This goal can be achieved by harnessing various data channels (manufacturers and distribution networks, online shops, large retailers, social media, market observers, call centers, press/magazines etc.) that retailers can leverage in order to gain more insight about potential buyers, and on the industry trends as a whole. This can enable the creation of novel on-line shopping experiences, the detection of influencers, and the prediction of upcoming fashion trends.

In this paper, we provide an overview of the main research challenges and an analysis of the most promising technological solutions that we are investigating in the FashionBrain project.

## 1 INTRODUCTION

In the world of fashion, retailers often either do not own enough data to predict customers' next trends or this data is not integrated in a way that can create valuable insights. Very few business platforms, e.g., search engines and social networks, have the required information about potential buyers, current and emerging trends, etc., and this increases retailers' dependence vis-à-vis these platforms. This dependence is aggravated by the fact that data is automatically accumulated to considerable extent by the rich repertoire of their user interactions and by the large investments search engines and social networks are doing in artificial intelligence. Talks with several CXOs confirmed that most existing fashion retailers in the mid 2000s decided not to invest massively in owning search engines but to pay global web search engines large amounts for SEO and advertisements. In order to alleviate the existing dependence from social networks and search engines, fashion retailers should be able to use their own tools and data to predict next emerging trends, and to acquire fashion related data by other means, for example by crowdsourced activities or by tailored user interactions.

In this paper we highlight the current data integration challenges in the fashion industry and present a vision of big data solutions that can provide a new data-driven fashion universe also enabling end-user applications like, for example, product search by image and fashion trend prediction that we are building in the context of the FashionBrain project[1].

Figure 1 illustrates the current and envisioned acquisition channels for an online-fashion shop. The current estimates in the figure may depend on various factors, such as the brand-strength, the niche-factor or the local shopping culture. Most traffic of the online shop is currently forwarded from major search engines. The shop owner 'buys' this traffic from search engine owners with advertisements displayed on the search engine, or with investments in search engine optimizations (SEO) of their own portal. Much fewer potential customers enter the shop directly navigating to the shop website and very few customers enter the shop via links from third party web pages, such as a blog from a fashion influencer.

Most retailers allow customers to shop via their built-in search engines, often still based on Lucene/Elasticsearch, a technology that implements text-retrieval algorithms from the late 70s to the early 90s. However, due to limited search effectiveness and visibility, the vast majority of customers falls back to a general purpose web search engine or to recommendations from social networks. The retailers can buy additional traffic from search engines or social networks at a great cost. In some cases the use of fashion bots [9] can help to interact with the social networks.

---

[1] http://fashionbrain-project.eu/

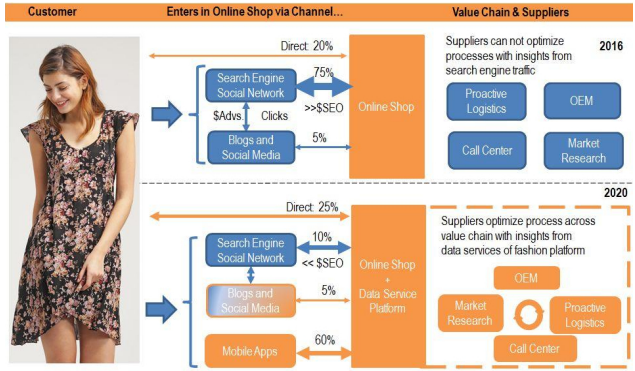

Figure 1: Typical and envisioned acquisition channels in online shops.

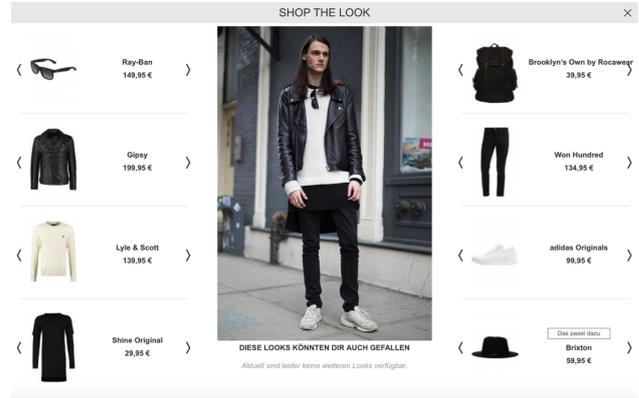

Figure 2: Example of photo-based product search with catalog matching and recommendation.

Another core objective for modern online fashion retailers is the predicting of new trends by analyzing social influencer on the web by machine-reading publicly available customer communication from text and image. Unfortunately, most retailers only rely on transactional data history of already bought items.

Moreover, the exclusive and insightful data about 'why the customer has entered the shop' is not owned by the shop-owner and can not be used by them for optimizing production, logistics, call center or market research activities.

To overcome the current limitations, we are working on an integrated approach to data management, data processing and predictive analytics, as explained in detail in the next sections. Such integrated approach will significantly improve the retailer capabilities of trend prediction and social media analysis (Section 4), which will in turn enable an advanced search by image capability with integrated product recommendation, as sketched in Figure 2 (Section 4).

### Use Case - Shop the Look

On top of our integrated fashion data platform, we envision end-user applications like the one presented in Figure 2 where the user is issuing a query consisting of an image (e.g., taken with a mobile device). The proposed system takes the input image and, thanks to a mix of image processing algorithms and crowdsourcing workflows, performs extraction of fashion products from the image followed by entity linking against a retailer product catalog that allows to disambiguate similar products. This allows the user to identify the exact product depicted in the image and enables him to purchase the product directly. More than search, the system recommends additional products (e.g., an hat) that is not depicted in the image but fits well with the identified products.

### Use Case - Fashion Trend Prediction

Using similar hybrid human-machine approaches [3] used for the previous use case where we combine image processing algorithms with human computation by means of crowdsourcing, we envision a system that takes as input Instagram images from a list of well know fashion bloggers. These data will be then processed over our pipeline that links depicted fashion products to retailer product catalogs. Then, thanks to time series analysis techniques that embed social media with sales data, the system will be able to *predict upcoming fashion trends*. Such predictions will enable retailers to better manage the supply chain and make sure that product stock is always up to date with expected demand.

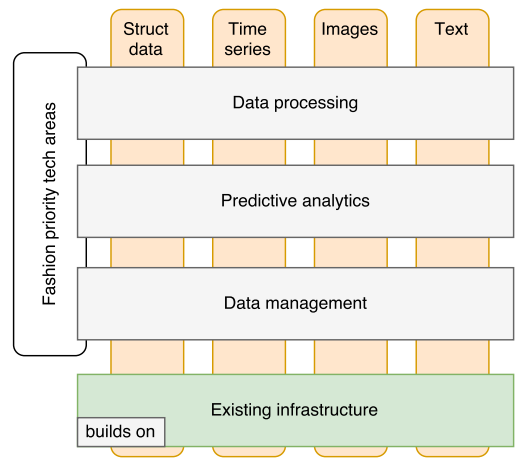

Figure 3: Fashion Big Data reference model.

## 2 CHALLENGES

As summarized in Figure 3, we seek an integrated approach for textual, structured, multimedia and longitudinal data. We focus on the following three core topics:

**Data processing.** The collection of heterogeneous data, its integration and curation is a problem broadly studied in the literature [6]. However, in the world of online fashion retail, typical high performance ad-hoc solutions are required. The main problems in this area are the integration amongst different data infrastructures and sources (e.g., from retailers, manufacturers, social media, logistics, website, customer care, etc.) and the complexity of the workflow needed to enable complex queries over available integrated data.

**Predictive analytics.** Predictive analytics in Big Data is also well studied [11]. In the world of fashion the main challenges we identified are the scalability of predictions and the lack of training data to build supervised models.

**Data management.** The practical implementation of high performance, state-of-the-art data management solutions is a challenge in the world of Big Data: the main direction we explore in our project is the development of low level, in-databases functionalities in the context of memory resident databases.

Next, we analyze more in detail how a successful online fashion retailer can improve each of these areas in the following section.

# 3 DATA PROCESSING

In Figure 4 our data processing workflow is sketched. We describe each step of the workflow in the following.

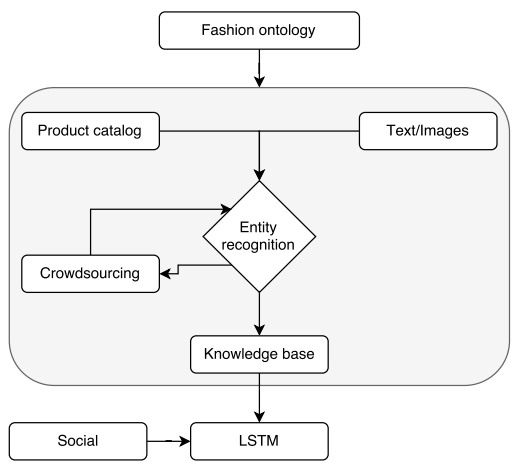

Figure 4: Data processing workflow.

## 3.1 Taxonomy Building

To support data integration across sources and data types and to have a central data schema, we propose to build a new fashion taxonomy based on redundancy-minimizing shopping categories (unlike the traditional fashion taxonomies on which e.g. clothes are divided depending on the gender). Moreover, the obtained hierarchical structure will be fine grained at a level that allows any item from any dataset to be linked to some layers of the hierarchy and subsequently to link fashion entities to a certain layer in the hierarchy. As a result, ee will be able to classify each item to the most appropriate category according to its type (clothing, shoes, accessory, etc.), its color and material.

This is a pivotal task that will help to obtain a more fine-grained data integration and a richer search experience.

## 3.2 Entity Linking from Text and Images

One of the main technological development to achieve the vision use cases described in Section 1 is *entity linking* from text and from images. The goal is to be able to identify and disambiguate fashion products against a product catalog.

Effective information extraction from unstructured content e.g., twitter, blogs, as well as multimedia sources such as Instagram, is critical for trends prediction. We will focus on the entity extraction process from text and image and mapping them as instances in our developed taxonomy from the rest of the workflow. We will extend our work on entity linking and instance matching [4] and entity recognition in idiosyncratic domains [10] with multilingual support and tune it to the fashion domain lexicon.

## 3.3 Crowdsourcing and Human Computation

To solve the problem of the lack of training data we make use of crowdsourcing. Example current uses of crowdsourcing for fashion data include the entity linking from images to product catalogs (see two use cases in Section 1 and processing of product reviews (e.g., extraction and classification of sizing issue mentions).

## 3.4 Relation Extraction and Exploitation Integration with Deep Learning

Retailers manage data about fashion products and transactions in a fashion data warehouse (FDWH), which is often a relational database management system. Recombining relational data from a FDWH with text data from blogs is therefore an important operation for learning about their users, monitoring trends and predicting new brands.

Our goal is to extend INDREX [8], an In-Database Text Mining System, with user defined table generating functions for joining relational data (entities and relationships) to text data and vice versa. Developers of fashion data-services will then be able to sample data and train models from a single, homogeneous data reservoir with standard SQL techniques and by a standard database connectivity layers, such as JDBC. Thereby, INDREX will leverage built-in optimizations for parallel query execution, column-based layouts and security features of the underlying MonetDB database.

We will learn a text join function for combining fashion entities in text data with entities from the fashion data warehouse. This includes (1) a robust entity mention detection function for recognizing idiosyncratic entity candidates in text, and (2) a join function from the FDWH to entity candidates in blogs. Moreover, we will integrate models used for linking entities in fashion blogs to the FDWH. Our methodology will base on Long-Short-Term Memory Networks (LSTMs) that have demonstrated excellent generalization capabilities, even for sparse and idiosyncratic data [5]. The overall results are an SQL-based text join operator, trained models for the fashion domain and a robust methodology that is applicable to other retail domains.

Text data might complement existing facts and dimensions in the FDWH. Given techniques of Open Information Extraction [1], we will learn n-ary relation extraction function for fashion related events. Our methods will execute the process of recognizing relations and entities jointly and will leverage LSTMs and other recurrent neural networks.

The fashion domain has a high turnover; often brands and products might appear every two months. Deep learning based methods might generalize well, and can be stacked into each other and with shallow learners (such as CRF based learners). The result is a learning framework that benefits from high generalization capabilities of deep learning models for basic language specific text mining tasks (such as recognizing entities) into models for domain specific tasks, such as linking entities or recognizing relations. Moreover, deep learning provides an inexpensive core that can support interactive crowdsourcing workloads.

## 4 PREDICTIVE ANALYTICS

We propose to perform a hierarchical prediction of fashion trends based on the history of items sales and their correlation with other histories. A time series $X = (t_1, v_1), ..., (t_n, v_n)$ is a set of $n$ temporal values $v_i$ that are ordered with respect to the timestamps $t_i$. We assume time series with aligned timestamps (possibly after a preprocessing step). Studying fashion as time progress has a pivotal importance because features considered to be fashionable change over time. For this, we aim to study fashion evolution and predict new trends. By accounting for evolving fashion dynamics for feed-back in the form of purchase histories, we hope to build systems that are quantitatively helpful for estimating users' personalized rankings (i.e., assigning likely purchases higher ranks than non-purchases), which can then be harnessed for recommendation. Since we will be dealing with time series of long histories (several years), the proposed prediction technique should be scalable with the length and the number of time series at a time (linear run-time complexity). We will implement an incremental version of the algorithm proposed in [7], that seems a promising scalable approach to the problem.

We will integrate matrix factorization with on-the-fly clustering techniques (e.g. [2]) to take in account the problem of sparsity of the data.

## 5 DATA MANAGEMENT

To provide more efficient data management solutions that can support scalable data integration approaches we propose to incorporate our data analytics features as part of a database management system. This is made possible by MonetDB, a highly performant column-store-based multicore, in-memory optimized RDBMS. The proposed functionalities will be implemented as primitives inside the kernel of MonetDB to optimize execution efficiency.

### High Performance Database Management Systems

Main Memory Databases (MMDB) are fundamental for high performance data management, but even more important is the integration of MMDBs with INDREX: a system that combines relation extraction and further exploitation with SQL into one RDBMS-based system.

Transaction oriented RDBMSs come with a high overhead to ensure the correctness of concurrent transactions. For analytical applications, many transaction constraints can be relaxed, so that much lighter-weight transaction management schemes can be used. So, analytics oriented RDBMSs usually only pay a small overhead to transactions, which is also the case of MonetDB.

Hadoop, MapReduce, key-value stores and distributed (R)DBMSs are also often used for some data analytics, however they are only suitable for relatively simple analytics, e.g. search for words, or word count. when the analytics becomes more complex (like in entity linkage), those systems quickly suffer from huge performance degradations. that makes them unsuitable for the analytics required in the world of fashion.

The combination of column store, main memory optimisation and analytics oriented approach makes a system such as MonetDB much more suitable for high performance data management than traditional RDBMSs, which are row store + disk optimised + transaction oriented.

## 6 CONCLUSIONS

The main technical challenges existing in the world of online fashion retail are due to the lack of data integration amongst different infrastructures and sources, complexity of the data workflow, scalability, and lack of training data for supervised models.

In this paper, we presented our vision on how to tackle these challenges proposing a data integration ecosystem based on core primitives of in-memory databases, deep learning over text, and crowdsourcing. These enables end-user applications like, for example, search product by image and fashion trend prediction.

## Acknowledgments

This project has received funding from the European Union's Horizon 2020 research and innovation programme under grant agreement No 732328.